\documentclass[conference,onecolumn]{IEEEtran}
\IEEEoverridecommandlockouts
\usepackage{cite}
\usepackage{amsmath,amssymb,amsfonts}
\usepackage{graphicx, float}
\usepackage{textcomp}
\usepackage{xcolor}
\usepackage{graphics}

\usepackage{csquotes}
\usepackage{subcaption}
\usepackage{multicol}
\usepackage{lipsum}
\usepackage[bookmarks=false]{hyperref}
\usepackage{amsmath,amssymb}
\usepackage{amsfonts}
\usepackage{float}
\usepackage{wrapfig}
\def\BibTeX{{\rm B\kern-.05em{\sc i\kern-.025em b}\kern-.08em
    T\kern-.1667em\lower.7ex\hbox{E}\kern-.125emX}}
    
\usepackage{tabularx}

\usepackage{algorithm}
\usepackage{algpseudocode}
\algblock{Input}{EndInput}
\algnotext{EndInput}
\algblock{Output}{EndOutput}
\algnotext{EndOutput}
\usepackage{authblk}

\usepackage{tikz}
\usetikzlibrary{matrix}
\usepackage{xparse}
\usepackage{soul}
\ExplSyntaxOn
\NewDocumentCommand{\avercalc}{O{1}+m}{%
  \clist_set:Nn \l_tmpa_clist {#2}%
  \fp_zero:N \l_tmpa_fp
  \clist_map_inline:Nn  \l_tmpa_clist {
    \fp_add:Nn \l_tmpa_fp {##1}
  }
  \fp_eval:n { round(\l_tmpa_fp/\clist_count:N \l_tmpa_clist, #1)}
}
\ExplSyntaxOff

\usepackage{float} % Add this package in the preamble
\newcommand*{\affaddr}[1]{#1} % No op here. Customize it for different styles.
\newcommand*{\affmark}[1][*]{\textsuperscript{#1}}
\newcommand*{\email}[1]{\textit{#1}}

\usepackage[most]{tcolorbox}
\tcbuselibrary{listings}
\usepackage{listings}

\usepackage[scaled=0.85]{beramono}

\usepackage{makecell}

\begin{document}

%\title{Zero-Shot and Few-Shot Time Series Forecasting in IoT Systems Using Large Language Models \\}

\title{In-Context and Few-Shots Learning for Forecasting Time Series Data based on Large Language Models \\}

\author[]{}

\author{Saroj Gopali\affmark[1], Bipin Chhetri\affmark[1], Deepika Giri\affmark[3], Sima Siami-Namini\affmark[2], Akbar Siami Namin\affmark[1]\\
\affaddr{Department of Computer Science\affmark[1]}  
\affaddr{Texas Tech University\affmark[1]}
\affaddr{Advanced Academic Programs Science\affmark[2]}  \\
\affaddr{Johns Hopkins University\affmark[2]}
\affaddr{Cumberland University\affmark[3]}
\\
\email{gopali.saroj@gmail.com}, \email{\{bipin.chhetri, akbar.namin\IEEEauthorrefmark{1}\}}@ttu.edu, 
\email{dgiri25@students.cumberland.edu}, \email{ssiamin1@jhu.edu}
}

\maketitle

\begin{abstract}
%Time series data play an important role in many application domains such as finance, businesses, and cyber-physical systems, to name a few. Due to its critical role, it is important to analyze these types of data properly and fit models that better describe the temporal properties and then can be used for prediction. 
Existing data-driven approaches in modeling and predicting time series data include ARIMA (Autoregressive Integrated Moving Average), Transformer-based models, LSTM (Long Short-Term Memory) and TCN (Temporal Convolutional Network). These approaches, and in particular deep learning-based models such as LSTM and TCN, have shown great results in predicting time series data. With the advancement of leveraging pre-trained foundation models such as Large Language Models (LLMs) and more notably Google's recent foundation model for time series data, {\it TimesFM} (Time Series Foundation Model), it is of interest to investigate whether these foundation models have the capability of outperforming existing modeling approaches in analyzing and predicting time series data. 
%Recent developments in the foundation models such as large language models (LLMs) have demonstrated high potential in adapting to various analytical tasks that include time-series analysis and predictions. %These capabilities are especially useful in the Internet of Things (IoT) context where data availability is limited, the conditions under which it operates may vary, and the overall generalization is essential. 
This paper investigates the performance of using LLM models for time series data prediction. We investigate the in-context learning methodology in the training of LLM models that are specific to the underlying application domain. More specifically, the paper explores training LLMs through in-context, zero-shot and few-shot learning and forecasting time series data with OpenAI {\tt o4-mini} and Gemini 2.5 Flash Lite, as well as the recent Google's Transformer-based TimesFM, a time series-specific foundation model, along with two deep learning models, namely TCN and LSTM networks. %The paper compares the performance of these various modeling approaches to meaningfully draw a conclusion about a better adaptation of these models for time series prediction. evaluate all the approaches using the SWaT industrial sensor dataset in terms of predictive accuracy and computational efficiency. 
The findings indicate that TimesFM has the best overall performance with the lowest RMSE value (0.3023) and the competitive inference time (266 seconds). Furthermore, OpenAI's o4-mini also exhibits a good performance based on Zero Shot learning. 
%Traditional deep learning models (i.e, LSTM and TCN)  find it difficult to model nonlinear changes when they do not require significant training costs. 
These findings highlight pre-trained time series foundation models as a promising direction for real-time forecasting, enabling accurate and scalable deployment with minimal model adaptation.

\end{abstract}


\section{Introduction}

Recent advances in foundation models such as large language models (LLMs), transformer-based architectures, generative models, and other deep neural network frameworks have enabled the possibility of better modeling and predicting of time-series data in various application domains such as finance and the Internet of Things (IoT). These models have proven to be highly accurate and effective in broader analytical domains than just traditional forecasting. For example, LLMs and transformer variants have been applied effectively to financial time series \cite{yu2023temporal}, pattern recognition in high-dimensional streams \cite{miah2024sensor}  as well as vulnerability detection \cite{ shiri2024systematic}. %Furthermore, these models also have been applied in problems such as detecting latent correlations and structural changes that influence long-term temporal behavior. 
Advanced transformer models such as BERT \cite{devlin2019bert} have also been utilized to predict the consequences of cyberattacks in interconnected control environments \cite{sana2024securing}, while neural network architectures have demonstrated their capabilities for robust anomaly detection across various domains . All of these advancements collectively represent a substantial shift toward more ``{\it context-aware}'', data-driven intelligence that enhances time series forecasting performance in rapidly changing environments including finance, healthcare and IoT.
% \cite{kasabov2013dynamic}

Traditional modeling approaches such as Autoregressive Integrated Moving Average (ARIMA)\cite{shumway2017arima} along with deep learning-based modeling such as Long Short-Term Memory (LSTM) \cite{lstm1997}, Temporal  Convolutional Networks (TCN) \cite{bai2018empirical} as well as Convolutional Neural Network (CNN) \cite{o2015introduction} have long been considered standard for sequential data modeling and prediction of time series data. 
While offering accurate modeling, these approaches require domain-specific architectures, large datasets for training, and a significant amount of model tuning that often restrict their applications and transferability in real-world scenarios. In addition, data heterogeneity, constrained processing resources, and real-time limitations pose substantial obstacles to the adaptation of these models in practice. With the rise of digitized ecosystems that generate vast multivariate observational data, foundation models are uniquely positioned to improve time series forecasting.

Recent emerging large language modeling (LLM) technologies and approaches have demonstrated remarkable generalization properties through ``{\it in-context learning}'', including zero-shot and few-shot learning, which enable performing prediction tasks for data that are not seen during training with little or no new training \cite{brown2020language}. Such advances in enabling in-context learning can address existing limitations while achieving remarkable performance in pattern recognition, reasoning, and context understanding across modalities. This versatility reflects a possible paradigm shift in time series analysis and prediction, where ``{\it limited}'' available time series data can be re-structured as text prompts (i.e., inputs) that an LLM can directly understand and analyze. By leveraging in-context learning, and notably the few-shot and zero-shot paradigms, we can significantly improve pattern-recognition and sequence-reasoning capabilities and be able to perform accurate forecasting without explicit training (or re-training) of any model on temporal context or task-specific fine-tuning \cite{liu2024lstprompt}.This adaptability is particularly useful in time series forecasting where system parameters change quickly. 

Inspired by these advances in LLMs, this paper investigates the degree of performance of in-context learning for time series prediction and analysis. As an application to investigate performance, the study uses the Secure Water Treatment (SWaT) dataset \cite{itrustdatasets} to explore the use of LLMs to predict through zero and few-shots in time series forecasting. In order to capture a wide range of modern architectures and inference strategies, we perform a systematic comparison on five models, including OpenAI o4-mini \cite{openai-systemcard},  Gemini 2.5 Flash Lite \cite{gelman2025gemini25}, TCN\cite{bai2018empirical}, LSTM \cite{lstm1997}, and Google TimesFM \cite{das2024decoder}. By comparing deep neural architectures, language-based models, and conventional time-series techniques, this study highlights the unique advantages, disadvantages, and implementation difficulties that arise in real-time forecasting. This paper makes the following key contributions:
\begin{enumerate}
    \item Investigates the performance of LLM-based ``in-context learning'' along with few/zero-shot learning methodologies in the context of time series prediction and analysis.
    \item Compares the performance and accuracy of prediction using LLM-based in-context learning with different foundation models such as TimesFM, OpenAI- o4-mini, Gemini 2.5 Flash Lite, and LSTM.
    \item Reports the results of outperforming LLM-based in-context learning compared to the state-of-the-art methodologies in time series analysis and prediction. 
\end{enumerate}

% google2025timesfm

The remainder of the paper is organized into the following sections. The related work is discussed in Section \ref{sec:related}. The technical overview of the models is presented in Section \ref{sec:overview}. Section \ref{sec:expriment} contains the information on dataset, the data preprocessing and the model configuration. Section \ref{sec:result} presents the results of the experiment. The limitation and discussion are presented in Section \ref{sec:diss}. Section \ref{sec:conclusion} concludes the paper with future work.

\section{Related work}
\label{sec:related}

\subsection{Deep Learning-based Approaches in Time Series Analysis}

Gopali et al.~\cite{9671488} assessed the effectiveness of Temporal Convolutional Networks (TCN) compared to Long Short-Term Memory (LSTM) networks for anomaly detection in time series, using comprehensive multivariate sensor datasets, including SWaT (Secure Water Treatment) and other recognized benchmarks for anomaly detection. The methodology involves a thorough analysis of temporal features, model architectures, and a comparative assessment across various evaluation metrics, including F1-score, recall, and precision. The findings demonstrate that TCN attains a maximum F1-score of 0.920, demonstrating enhanced training times and stability. In contrast, deeper LSTM models modestly improve precision but incur higher resource consumption and reduced recall. This research illustrates the stability and efficiency of TCNs, while highlighting the challenges of diminished interpretability relative to sequential RNNs and the substantial resource demands of deep LSTMs, which complicate their practical application for real-time anomaly detection in industrial settings.

Siami-Namini et al.\cite{8614252} conducted a systematic comparison of the ARIMA and LSTM models for time series forecasting, focusing on predicting values for several economic and financial datasets, including monthly closing prices from stock data and sequences with diverse lengths. The author's work outlines the steps necessary for achieving stationarity in ARIMA models and specifies the design parameters for LSTM, including the optimization of hyperparameters. In addition, the study indicates that the number of training epochs does not influence the performance of the forecasting model, which operates predominantly in a random manner. The results of the experiment indicate that LSTM networks achieve an average error reduction of 84–87\% compared to ARIMA, particularly for datasets exhibiting nonlinear or chaotic patterns, demonstrating the superior ability of the deep learning model.
However, LSTM models depend on larger training samples and precise hyperparameter tuning for effective generalization. In addition, they may be computationally demanding when applied to smaller datasets, leading to potential instability or overfitting.

Similarly, Gopali \& Namin ~\cite{electronics11193205} tested Bidirectional LSTM, LSTM, CNN-TCN, and CuDNN-LSTM in large datasets, including SWaT and industrial IoT sensor logs, to investigate a variety of deep learning-based methods for time series anomaly detection in Internet of Things (IoT) environments. Their empirical findings show that while TCN is noticeably faster and more scalable to larger datasets, CuDNN-LSTM outperforms others in achieving the highest detection accuracy. CuDNN-LSTM model performs better than other models when using different timestamps (i.e., 15, 20, and 30 min), while the TCN-based model is trained more quickly. The authors highlight that the computational complexity, memory requirements, and the requirement for large amounts of labeled training data hinder the deployment of these deep learning solutions for low-power or real-time IoT devices, despite their advantages in detection capability.

Siami-Namini et al.\cite{BiLSTMLSTM} have examined performance of the LSTM and BI-LSTM models on financial time-series forecasting by using the historical daily, weekly, and monthly adjusted-close price of some major market indices and blue-chip stock, such as the S\&P 500, NASDAQ, Dow Jones, Nikkei 225, Hang Seng Index and IBM. Their design used rolling one-step-ahead forecasting, and both LSTM and BI-LSTM were trained using the same setups, using RMSE as the evaluation metric. It was indicated that BI-LSTM was always higher in performance as compared to standard LSTM with an average of 37.78 \% decline in the forecasting error.  The study revealed that the benefit of temporal pattern learning in both forward and backward directions was realized. The paper demonstrates the usefulness of bidirectional recurrent networks to the financial prediction problem and offers a better benchmark than the classical LSTM networks.

% Review Our own papers I shared with you through email here. 

% The paper related to LLM-based approach to code generation for LSTM code can be reviewed in the next subsection. 

\subsection{LLM-based Approaches in Time Series Analysis}
Parker et al.\ \cite{parker2025eliciting} introduced reinforcement-learning frameworks called COUNTS that encourage chain-of-thought (CoT) reasoning in LLMs to analyze numerical time-series data. The COUNTS framework was evaluated on three different benchmarks: (1) ECG-QA for medical Time Series reasoning, (2) Context-Is-Key (CiK) for contextual forecasting across multiple domains, and (3) the UCR Time Series Classification Archive for few-shot classification. COUNTS significantly improves performance in medical signal analysis, contextual prediction, and few-shot classification tasks by integrating high-fidelity discrete tokenization, supervised fine-tuning on a variety of time series tasks, and reinforcement learning. COUNTS outperformed Gemini 2.5 Flash and OPENAI o4-mini with 66.5\% in ECG-QA, 54.5\% SMAPE for CiK forecasting and 60.1\% accuracy on UCR. However, COUNTS mainly focuses on univariate and domain-specific time series data, making it less suitable for multivariate time series forecasting tasks.

Mapato et al.\ \cite{mapato2025evaluating} conducted a comparative analysis among Vision Language Models (VLMs) and Optical Character Recognition (OCR) based on Retrieval-Augmented Generation (RAG) to evaluate 13 models on Thai academic documents. Their experiments utilized a golden dataset of 18 pages that comprised six types of document (abstract, equation, table, and main content) and were evaluated considering character error rate (CER), word error rate (WER), and retrieval accuracy. The results showed that Gemini-2.Flash (15.5\%) achieved the lowest CER and OpenAI o4-mini attained the highest retrieval accuracy (98.9\%) although the CER was around 37.1\%. 

Cao and Wang \cite{cao2024evaluation} assessed the LLMTIME framework, which employs GPT-3.5-Turbo-Instruct 
for zero-shot forecasting by converting numerical sequences into text tokens, and compared it with the traditional ARIMA 
statistical model. Their experiments utilized statistical benchmark datasets such as Darts  Monash and synthetic almost-periodic signals sourced from CEIC data . The results showed that ARIMA consistently achieved a lower Mean Squared Error (MSE) across all datasets, i.e AirPassengers (ARIMA = 2441.7 
vs. LLMTIME = 10380.6) and FredMd (ARIMA = 574,689 vs. LLMTIME = 11,344,708) which demonstrates its robustness on 
structured, stationary data. However, LLMTIME token-based textual representation struggled to capture non-linear 
temporal dependencies and multi-feature interactions, resulting in poor performance on noisy signals. 
% \cite{herzen2022darts}, \cite{godahewa2021monash},  \cite{ceicdata_2025}

Jin et al.\cite{jin2023time} propose the TIME-LLM framework that adapts pre-trained, frozen large language models such as LLaMA-7B  and GPT-2 for time series forecasting through a reprogramming-based  approach. The key innovation lies in transforming temporal sequences  into text prototype representations and utilizing the Prompt-as-Prefix (PaP) technique to bridge the modality gap between numerical time series and natural language tokens. Extensive experiments covering short- and long-term prediction tasks were carried out on popular benchmark datasets such as ETTh1/ETTh2, ETTm1/ETTm2, Weather, Electricity, Traffic and ILI. The authors present the observed average performance improvements of 12\% compared to GPT4TS and 20\% compared to TimesNet, respectively.  TIME-LLM exhibits strong few-shot and zero-shot forecasting capabilities, with less than 0.2\% of LLaMA-7B's the total parameters used for adaptation. However, the effectiveness of this method still relies on the existence of semantic alignments between text prototypes and underlying time series patterns, which may limit its generalizability in domains where such correspondence is absent.
% TIME-LLM demonstrated state-of-the-art accuracy when compared to widely used statistical and transformer-based baselines that include PatchTST, FEDformer, and DLinear.
% \cite{zhou2021informer},\cite{wu2023timesnet},\cite{makridakis2000m3},\cite{makridakis2018m4}

Gopali et al. \cite{GOPALI2024100120} examine how mainstream LLMs generation in time-series forecasting and test the generated LSTM models based on LLM models with prompt-engineering approaches. The authors utilized LLM models such as ChatGPT, PaLM, LLaMA 2 and Falcon to perform code generation. The generated code with the textual criteria of prompt-engineering such as clarity, objective, richness of context, and formatting. The finding shows that the models generated by the LLM  in some cases are comparable to manually crafted models. Their results indicate that the GPT 3.5  models reached an RMSE of 0.035, while the manual models reached an RMSE of 0.033. On the contrary, LLaMA-2 and Falcon produced models that had more errors, and RMSE levels increased to 0.065-0.081. This paper has demonstrated the potential and limitations of LLM-based model generation by demonstrating that although GPT 3.5 is capable of closely replicating the code written by experts. Its model quality differs greatly depending on the family of LLMs and prompt conditions.

Ekambaram et al. \cite{ekambaram2024tiny} designed the Tiny Time Mixer (TTM) framework to overcome the scalability challenges posed by large transformer-based models. The proposed framework offers a lightweight temporal-mixing backbone enhanced by adaptive patching and resolution prefix tuning. Three distinct variants, i.e. TTM-Base (1M), TTM-Enhanced (4M), and TTM-Advanced (5M), were pre-trained on a substantial dataset of one billion samples, utilizing six A100 GPUs. Comprehensive evaluations were performed on the ETT \cite{zhou2021informer}, Electricity, Traffic and Weather datasets. TTMs outperformed leading models such as Moirai, Chronos, TimesFM, and GPT4TS while achieving up to 40\% lower MSE. In few-shot scenarios with only 5\% of the data, the TTMB variant exceeded GPT4TS and Time-LLM by 15\% and 10\%, respectively. While these results are promising, TTM only focuses on point forecasting with fixed context lengths, needing work in probabilistic prediction and multi-task adaptation.

\section{Technical Background}
\label{sec:overview}
%This section provides the preliminary of models and schemas used in this study. 

\subsection{The OpenAI o4-mini}
OpenAI o4-mini was introduced in 2025 \cite{openai-systemcard}, prioritizing cost-efficient reasoning and multi-modal capabilities that support both text and vision. o4-mini performs at much less computational cost and latency, which is suitable for being applied in real-time and resource limited scenarios and would work best in time series data. When compared to current models, o4-mini demonstrated impressive accuracy with the American Invitational Mathematics Examination (AIME) \cite{opencompassAIME2025} 2025 dataset score of 92.7\% . It even has its own reasoning abilities against larger models, making it a solid choice for embedded forecasting tasks. System-level analysis indicates that o4-mini maintains strong chain-of-thought reasoning capabilities and bears the least bias, even with compact architecture. OpenAI's o4-mini, which is used for reasoning tasks, achieved a decent 72.6\% Symmetric Mean Absolute Percentage Error (SMAPE)\cite{parker2025eliciting} in textual forecasting, which outperforms the traditional LLMs.
%   \cite{datacamp_o4mini_2025}

\subsection{Gemini 2.5 Flash lite}
Gemini 2.5 Flash Lite \cite{gelman2025gemini25} released in 2025 is a lightweight advanced large language model 
designed for fast and cost-effective reasoning. By supporting up to one million tokens and offering multi-modal 
capabilities \cite{comanici2025gemini}, including both textual and visual inputs, Gemini 2.5 Flash Lite has 
demonstrated robust applicability for embedded systems and Internet of Things (IoT) platforms. Flash Lite maintains competitive accuracy in complex reasoning and time-series
 forecasting tasks at significantly lower computational costs while achieving inference speeds that 
are roughly 1.5 times faster than its predecessor. The model architecture is appropriate for time series and anomaly detection, since it enables adaptive processing and integration of real-time tools. All of these 
qualities make Gemini 2.5 Flash Lite a viable platform for our research, as well as for resource-efficient forecasting 
and analytics in time series forecasting. 
% According to \cite{deepmind2025flashlite},

\subsection{The Google TimesFM}

TimesFM  \cite{das2024decoder} is the foundation model for time-series forecasting that uses a transformer-based decoder-only architecture. It was trained on a wide range of large-scale chronological datasets. TimesFM is pre-trained on a huge time-series corpus comprising 100 billion real-world points. Despite being significantly smaller (200M parameters) than the most recent LLMs, its zero-shot performance is impressive on a range of unseen datasets of various domains and temporal granularity. The system models local and global patterns across several domains, and textual information obtained from multiple sources, such as open-source code, such as common crawl, enables TimesFM to identify language trends. For example, the model can summarize and respond to queries about current events when combined with retrieval. Our forecasting benchmarks' goals are directly aligned with TimesFM capacity to generalize across invisible domains and temporal granularity.

Das et al.\ \cite{das2024decoder} provide the detailed architecture and data set used to train the model. TimeFM is a decoder-only transformer architecture that takes in continuous time series data divided into non-overlapping patches for input and processed by a residual multi-layer perceptron (MLP) block that generate patch embedding. After being combined with learned positional encodings, these embeddings are run through a stack of 20 transformer layers (for the 200M parameter model), each of which has feedforward sublayers and multi-head self-attention (16 heads), which is followed by a normalization layer. With this setup, the model can effectively capture both short- and long-range dependencies. In addition, the architecture makes use of causal masking to guaranty autoregressive forecasting and rotary positional embeddings (RoPE) to maintain natural ordering of temporal input during attention computation.

A corpus of over 100 billion time points combining both synthetic and real-world data sources was made available to TimesFM for pretraining \cite{das2024decoder}. Crucial foundation datasets include Wikimedia Pageviews\cite{wikimedia2023} (with hundreds of billions of records from 2012–2023), Google Trends \cite{googletrends2023} (with hourly, daily, weekly, and monthly interest for approximately 22,000 queries from 2007–2022), and canonical forecasting benchmarks such as M4 , Electricity, Traffic, and ETT . Each batch of pretraining comprises 20\% synthetic and 80\% real data, equally weighted across various temporal granularities. In order to minimize the mean squared error (MSE) over anticipated future patches, optimization is carried out using AdamW with cosine learning rate decay. The model uses dropout regularization, large-batch distributed optimization across multiple accelerators, and mixed-precision computation to improve training stability and convergence from diverse data sources.
% \cite{wikimedia2023} \cite{makridakis2022m4} \cite{googletrends2023} \cite{zhou2021informer}

\subsection{Temporal Convolutional Networks (TCN)} 
Temporal Convolutional Networks (TCNs) \cite{lea2017temporal} are a type of deep learning model that is suitable for time series data. Unlike traditional CNNs built for spatial inputs like images, TCNs are designed to handle sequential or temporal data. TCN is used in machine learning in sequential data comparison, such as time series and speech recognition. TCN combines causal filters with dilated convolutions and this makes the network more efficient in the processing of lengthy data sequences. This is achieved through an expansion of the field length of the receptive network to the receptive network field length of the network. TCN uses causal filters which do not depend on future inputs in order to give output, and this makes the network more efficient. The residual connections used in TCN are also used to train the deeper layers of the network.

\subsection{Long Short-Term Memory (LSTM)}

Long Short-Term Memory (LSTM) networks \cite{lstm1997} are a particular type of recurrent neural networks (RNNs) that are intended to represent long-term time dependencies in sequence data. They are particularly useful in tasks such as time series prediction in which the ability to store information over a long period of time is required. In contrast to regular RNNs, LSTMs have a collection of memory cells that is controlled by three gating mechanisms, namely the input gate, the output gate, and the forget gate. 
% These gates regulate the ways information is stored, updated, or retrieved in the memory cells and this allows the model to deal more with the long-term contextual cues. 
The input gate is used to decide what sort of new information can affect the cell state and the output gate is used to decide what information can be revealed to the next layer or prediction task. The forget gate determines what is to be forgotten in a given time period based on the information that has been previously stored. 
% This gate has the potential of effectively erasing the old material by scaling certain aspects of the cell state to zero but the values that are near one show that the information is required to be retained.

\subsection{In-Context Learning}

The capability of large language models to use the input prompt to condition new tasks without performing any parameter updates is known as in-context learning (ICL). ICL introduced in GPT-3 \cite{brown2020language}, and allows models to generalize through a task by implicitly learning task structure using a few demonstrations. Recent research has demonstrated that ICL is the result of internal processes that are similar to meta-learning in which models are trained to learn to map input-output correlations by paying attention to patterns, instead of relying on gradient-based optimization \cite{xie2021explanation}. In addition, there is research that indicates that transformer architectures execute a type of in-context gradient descent, allowing those to estimate learning algorithms internally during inference \cite{von2023transformers}. These results place ICL as an important ability to perform adaptive tasks, quick generalization, and adaptive reasoning in contemporary large language models.

\subsection{Zero Shot Learning}
Zero shot learning is an approach where models are required to generalize entirely new tasks or domains without being exposed to task-specific examples during training. Its roots are based on the development of large language models and transformer architectures \cite{vaswani2017attention}, which enables models to capture complex relationships and long-range context using self-attention mechanisms. The first demonstrations of zero-shot reasoning in language models emerged with the application of transformer-based LLMs for text and were later extended to structured and time-series data by reformatting temporal signals into textual or symbolic tokens \cite{zhang2024survey,sun2024symbolic}. Recent surveys confirm that these methods, which treat time series as a language, achieve competitive accuracy against established traditional approaches \cite{zhang2024survey}. This flexibility and transferability have made zero-shot models especially applicable in the IoT and industrial contexts, where continuous adaptation and limited annotation are major practical considerations.

\subsection{Few Shots Learning}
few-shot learning expands on the zero shot paradigm by allowing models to condition on a small number of annotated instances from the target domain, allowing more precise adaptation and calibration to new tasks \cite{brown2020language}. In practical terms, this involves transformer-based LLMs with several example input-output pairs as context often called ``in-context'' learning, which the model uses to infer the underlying task structure before generating new predictions \cite{sun2024symbolic}. Recent methodological advances have revealed that even minimally supervised demonstration (a few-shot promptings) is sufficient for large models to exceed the performance of common shallow baselines in forecasting and classification tasks over time-series data \cite{zhang2024survey}. This opens possibilities for deploying high-performing models in environments where data labeling is expensive or rapid system changes are frequent, such as in time series data.

\section{Experimental Procedure}
\label{sec:expriment}

\subsection{Dataset}
In the experiment, we used a subset of the SWaT dataset (July version 2, 2019 dataset) of iTrust, Center for Research in Cyber Security, Singapore University of Technology and Design \cite{itrustdatasets}. The information in this data set is taken into the Secure Water Treatment (SWaT), which is an experimental water treatment that cybersecurity researchers utilize to design safe Cyber Physical Systems. The data is related to 3 hours of SWaT running under standard operating conditions and 1 hour of SWaT running at attack incidents. The feature set of this data set will include network traffic and values of sensors and actuators that are identified by their normal and abnormal behaviors. The dataset contains 14996 data points with 78 columns.

\subsection{Data Reprocessing}
In the preprocessing stage, missing values in the target variable (\texttt{LIT 301}) were removed to ensure data consistency. The target series that reflects the sensor reading of the water tank level was changed to a floating-point numerical array to be modeled. A sliding window method was used to exploit the temporal dependencies, in which the window size of \texttt{720} time steps was meant to compose input sequences and the next single step horizon was taken as the prediction target. This conversion generated overlapping time series windows, making the model able to learn sequential patterns. To avoid model bias, the data were then split into training \texttt{70\%}, validation \texttt{15\%}, and testing \texttt{15\%} subsets. A \texttt{StandardScaler} was used to normalize all features fitted in the training data and applied to the validation and test data to avoid data leakage. This step of standardization guaranteed the numerical stability and enhanced the convergence of the model optimization. The data preprocessing stage was implemented in Python using the Pandas and NumPy libraries, and training was performed using PyTorch with GPU acceleration when available.

\subsection{Model Configuration}

\begin{figure*}[t]
    \centering
    \includegraphics[width=\linewidth, height=8cm]{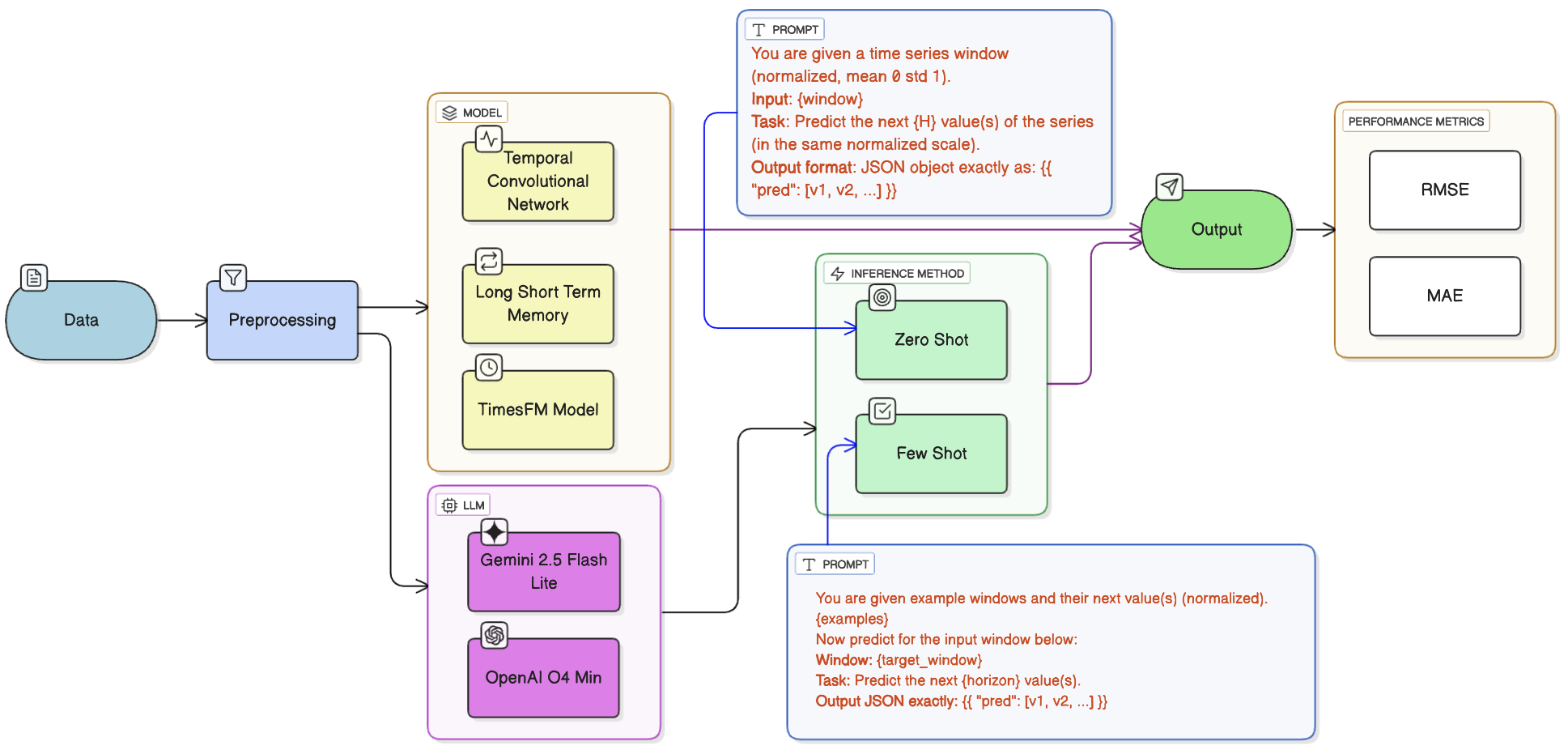}
    \caption{Time-series forecasting workflow pipeline.}
    \label{fig:llm_flow}
\end{figure*}

Figure \ref{fig:llm_flow} illustrates the forecasting workflow that is being implemented in the study, which incorporates both deep learning and large language model inference designs. The SWaT data are first subjected to the operations needed to preprocess the data prior to being sent into two modeling tracks: (i) traditional deep learning models and Transformer based Model: (1) TCN, (2) LSTM, and (3) TimesFM. (ii) parameter-efficient foundation models: (1) Gemini 2.5 Flash Lite, (2) OpenAI o4-mini . In the case of the LLM-based methods, the data windows are transformed into structured natural-language prompts with the specification of the prediction horizon and output format. 

Two inference strategies are considered. In zero-shot prompting, when the model is fed with the target input window only. In addition to zero-shot, few-shots prompting provides example windows and the corresponding outputs to the model with guidance on identifying temporal patterns.

\begin{table}
    \centering
\caption{Summary of Deep Learning Models Architectures.}
    \begin{tabularx}{\linewidth}{ll}
    
        \hline
        \textbf{Model} &\textbf{Layers} \\ \hline

        TCN & \makecell[l]{- 3TCN layers (unit = 64, dilations = [1,2, 4])\\
                           - Dropout (0.2)} \\ \hline
         LSTM &  \textit{- 2 LSTM layers (unit = 64)}    \\ \hline
 
    \end{tabularx}

    \label{tab:dl_arch}
    % \vspace*{-0.15in}
\end{table}

Furthermore, each of the deep learning models (i.e., LSTM and TCN) was trained with equal hyperparameter values to allow a fair comparison between the supported architectures, as shown in Table \ref{tab:dl_arch}. The deep learning models were trained with \texttt{20} epochs and a batch size of \texttt{32} which can support mini-batches and do not lose context over time. The optimization was performed using Adam optimizer with a learning rate of \texttt{$1\cdot10^{-3}$}.

In this experiment, TimesFM is utilized in an inference-only mode, i.e. the pretrained model is run on the normalized test windows without further training or fine-tuning. Each input sequence is then transformed by the model into a single-point forecast by its built-in \texttt{forecast()} method, which takes the time context and returns the next value on the normalized scale. The main point of this approach is that TimesFM has an essential benefit in practical deployment, which is the model ability to make accurate and efficient forecasts with a minimum of computational costs. In this process, the model does not need gradient updates, extensive training, or hyperparameter optimization to demonstrate competitive performance.

Regardless of the model family, all forecasting outputs are evaluated using standard regression performance metrics, specifically the Root Mean Squared Error (RMSE) and Mean Absolute Error (MAE). This unified pipeline enables a consistent, controlled comparison between architecture-driven learning and prompt-guided generalization capabilities for time-series forecasting.

\section{Results}
\label{sec:result}

Table  \ref{tab:model_comparison} shows the performance comparison across models highlights a clear advantage for foundation model based time-series forecasting (i.e., TimesFM). TimesFM performs the best of all the models analyzed with the smallest RMSE ($0.3025$) and a competitive MAE ($0.2127$) with a relatively short inference time (266 seconds). These findings show that TimesFM has high predictive validity and efficiency and, therefore, can be used in volatility-sensitive prediction problems.

The OpenAI o4-mini model in zero-shot form is also competitive with the second-best RMSE of $0.3310$ and the lowest MAE of $0.2098$. Even though o4-mini is a very robust pre-trained model and it does not require task-specific examples, it is much more expensive to computation than TimesFM, taking over 21000 seconds. o4-mini in its few-shot variant achieves lower accuracy (RMSE 0.3727) and this could be an indication that it is sensitive to prompt structure or that few-shot conditioning is not as useful in the domain as it is in other ones.

Gemini 2.5 Flash-Lite has significantly poorer predictive accuracy both in zero-shot and few-shot (RMSE $\geqslant$ 0.4060), and takes longer time for prediction compared to TimesFM. Those findings indicate that the general-purpose foundation models do not necessarily transfer well to time-series prediction tasks, and on the need to adapt to the domain.

The models include TCN and LSTM which have RMSE of 0.7174 and 0.7361 have the highest RMSE respectively. The fact that they have a higher MAE and also slower convergence is an additional indication of their inability to fully capture the nonlinear volatility patterns that are evident in the data. This disparity in performance confirms that classical deep learning designs can perform poorly without substantial data sets, architecture engineering, or the synthesis of exogenous features.

\begin{table}[t]
    \centering

    \caption{Performance Comparison Between Models.}
    \label{tab:model_comparison}
    \scalebox{0.95}{
    \begin{tabular}{lrrr}
    \hline
    \textbf{Model} & \textbf{RMSE} & \textbf{MAE} & \textbf{Training Time (s)} \\ 
    \hline
    LSTM & 0.7361 & 0.6714 & 1,333.53 \\ 
    TCN  & 0.7174 & 0.6293 & 1691.67 \\ 
    \hline
    \textbf{Model} & \textbf{RMSE} & \textbf{MAE} & \textbf{Inference Time (s)} \\ 
    \hline
    TimesFm & {\bf 0.3025} & 0.2127 &266.0 \\
    o4-mini Zero Shot & 0.3310 & 0.2098 &21,306.65\\
    o4-mini Few Shot & 0.3727 & 0.2243 &32148.21\\
    Gemini 2.5 Flash-Lite Zero Shot & 0.4060 & 0.2242 &5,221.49\\
    Gemini 2.5 Flash-Lite Few Shot & 0.5134 & 0.3368 &11,110.16\\

    \hline
    \end{tabular}
    }
    \vspace{-0.3in}
\end{table}

These results underline the efficacy of contemporary, pretrained time-series foundation models. TimesFM is not only the most accurate but also exhibits better computational efficiency, thus it is the most feasible of all the methods tested in the real world forecasting environments.

% \begin{figure}
%     \centering
%     \includegraphics[width=\linewidth]{Figure/timesfm.png}
%     \caption{Comparison of Historical and Forecast Results Using the TimesFM Model}
%     \label{fig:timesfm_pred}
% \end{figure}

% \begin{figure}
%     \centering
%     \includegraphics[width=\linewidth]{Figure/tcn.png}
%     \caption{Comparison of Historical and Forecast Results Using the TCN Model}
%     \label{fig:tcn_pred}
% \end{figure}

% \begin{figure}
%     \centering
%     \includegraphics[width=\linewidth]{Figure/lstm.png}
%     \caption{Comparison of Historical and Forecast Results Using the LSTM Model}
%     \label{fig:lstm_pred}
% \end{figure}

\begin{figure*}[t]%[htb]   

     \centering

     \begin{subfigure}[b]{0.30\textwidth}
         \centering
         \includegraphics[width=\textwidth ,]{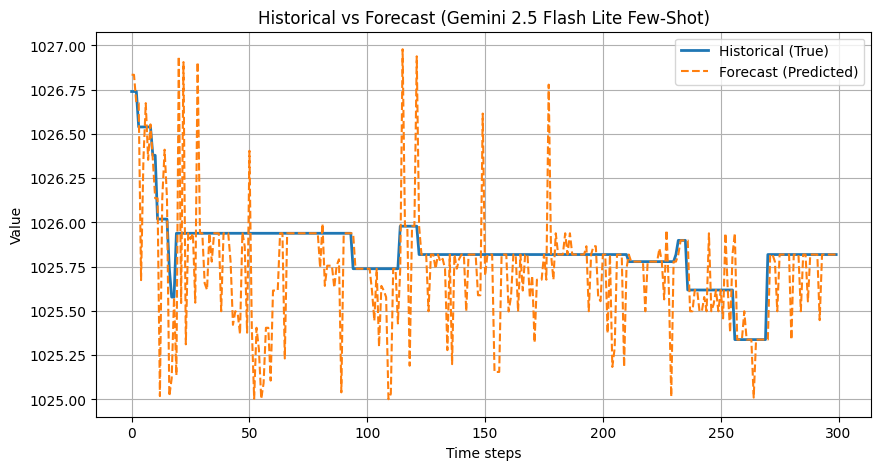}
         \caption{\footnotesize{Gemini 2.5 Flash lite Few Shot}}
         \label{fig:litefew}
     \end{subfigure}
     \hfill
     \begin{subfigure}[b]{0.30\textwidth}
         \centering
         \includegraphics[width=\textwidth]{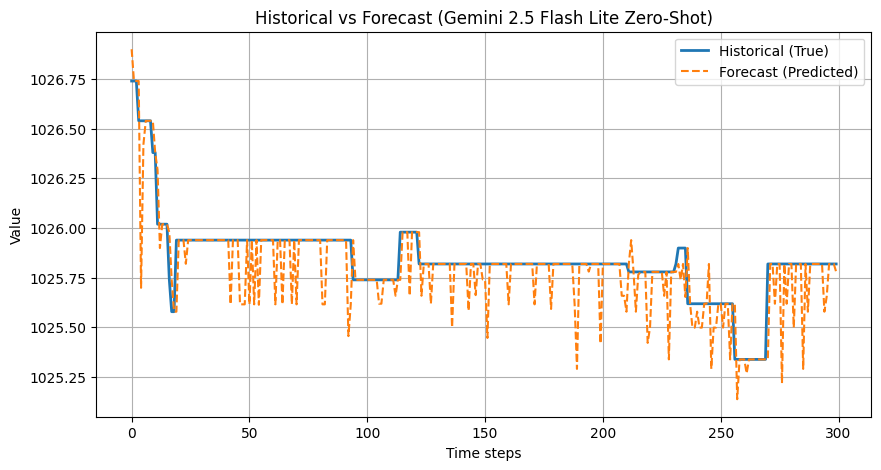}
         \caption{\footnotesize{Gemini 2.5 Flash lite Zero Shot}}
         \label{fig:litezero}
     \end{subfigure}
     \hfill
    \begin{subfigure}[b]{0.30\textwidth}
         \centering
         \includegraphics[width=\textwidth]{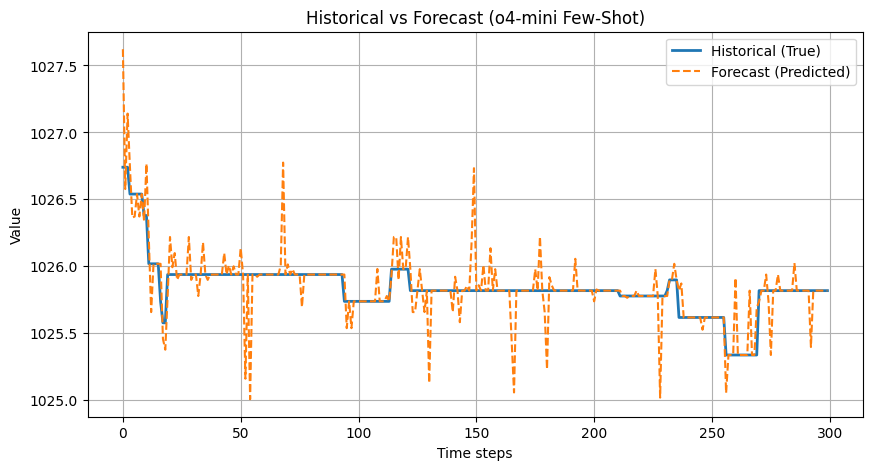}
         \caption{\footnotesize{ OpenAI o4-mini Few Shot}}
         \label{fig:o4few}
     \end{subfigure}
      \par\bigskip
      \hfill 
       
      \begin{subfigure}[b]{0.30\textwidth}
         \centering
         \includegraphics[width=\textwidth]{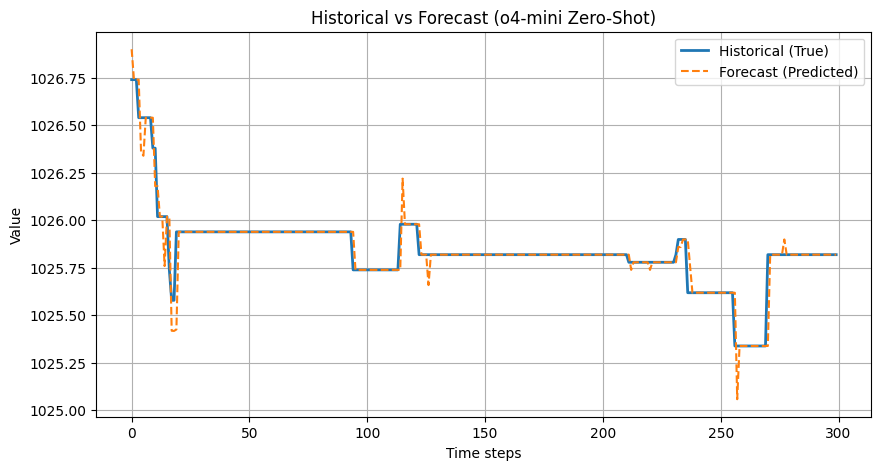}
         \caption{\footnotesize{OpenAI o4-mini Zero Shot}}
         \label{fig:o4zero}
     \end{subfigure}\hfill
     \begin{subfigure}[b]{0.30\textwidth}
         \centering
         \includegraphics[width=\textwidth]{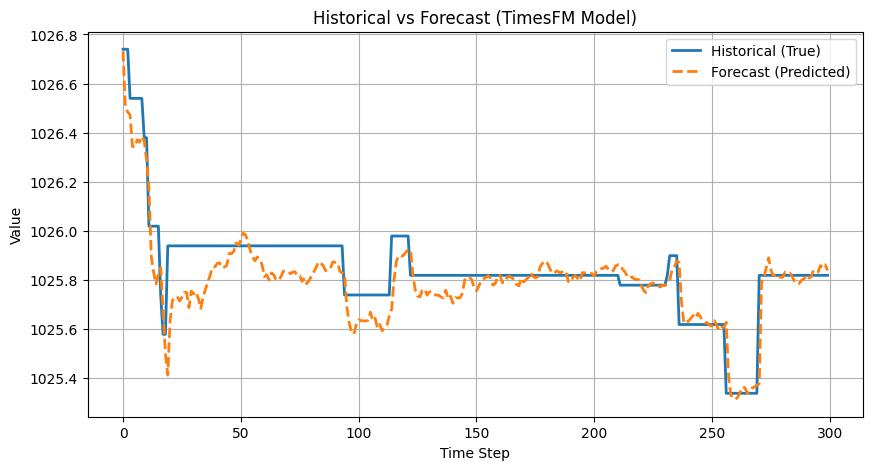}
         \caption{\footnotesize{TimesFM}}
         \label{fig:timesfm}
     \end{subfigure} \hfill
     \begin{subfigure}[b]{0.30\textwidth}
         \centering
         \includegraphics[width=\textwidth]{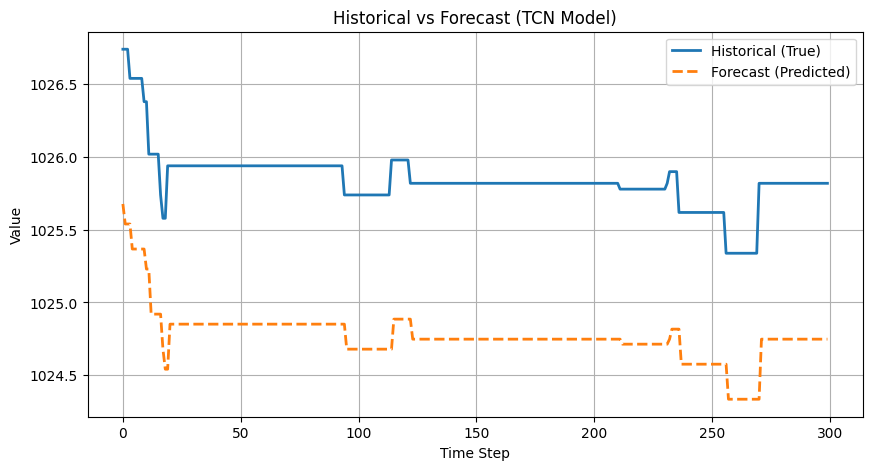}
         \caption{\footnotesize{TCN}}
         \label{fig:tcn}
     \end{subfigure}
     \begin{subfigure}[b]{0.30\textwidth}
         \centering
         \includegraphics[width=\textwidth]{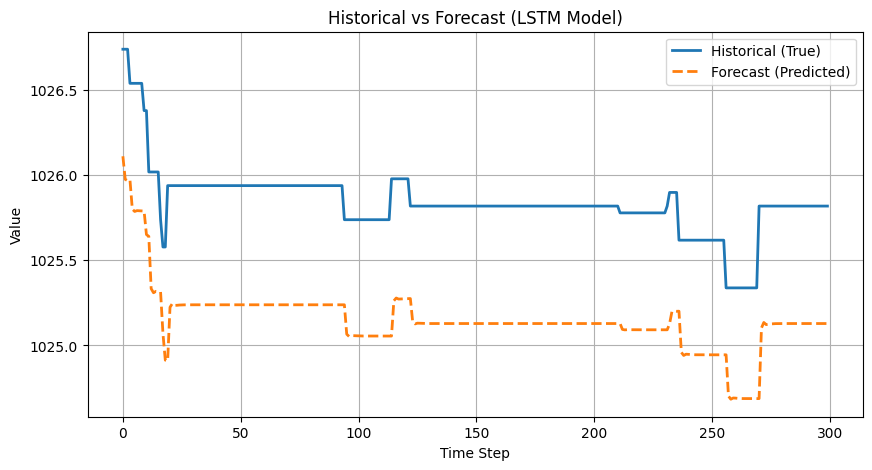}
         \caption{\footnotesize{LSTM}}
         \label{fig:lstm}
     \end{subfigure}
         
\caption{Comparison of Historical and Forecast Results of Models (First 300 data points).}
\label{fig:comp_fig}
\end{figure*}

Figure \ref{fig:comp_fig} provides a visual comparison of the first 300 data points from the historical SWaT and the one-step-ahead predictions of all seven tested models. The plots show the actuality of the predicted behavior of each model towards the actual process behavior in the initial part of the test set. TimesFM in Figure \ref{fig:timesfm} shows the best correspondence to the true signal, which allows us to present the rapid variations and follow them smoothly and accurately. Whereas OpenAI o4-mini in Figure \ref{fig:o4zero} in zero-shot mode is also inclined in the same direction with only slight deviations. In contrast, Gemini 2.5 Flash Lite in Figure \ref{fig:litefew} \ref{fig:litezero} and deep learning models (TCN in Figure \ref{fig:tcn} and LSTM in Figure \ref{fig:lstm}) have a higher lag (i.e., delay) and larger residual errors, which continue to grow with the horizon.

\section{Discussion}
\label{sec:diss}

\subsection{Performance Comparison}
The model comparison shows that pretrained foundation models have a clear edge in multivariate temporal dependencies of time series data. The low RMSE and high inference efficiency of the TimesFM decoder-only architecture and its high scale pre-training suggest that ``contextual reasoning'' in sequence prediction is a valuable component. The large-scale pre-training can effectively generalize to unseen sensor patterns without task-specific training. The zero-shot performance of o4-mini further demonstrates the effectiveness of contextual reasoning in time series sequence prediction, but its high computational latency limits its use in resource-constrained edge devices. Although designed to support lightweight reasoning, Gemini 2.5 Flash Lite, at the cost of accuracy, is slower than the individual forecasting capabilities of general-purpose language modeling. This suggests that purely nonlinear cyber-physical dynamics cannot be addressed with lightweight reasoning.

The significantly worse performance of conventional deep learning based architectures indicates the weaknesses of models that are based purely on local temporal context and need a large amount of supervised training with labeled data. Their reliance on feature engineering that is specific to a domain and long optimization periods is in stark contrast. The difference between this performance can be further increased when operating in dynamic time series environments where anomalies, sensor drift, and operational changes can often change the systems behavior. In general, the findings support the fact that the foundation models provide significant advantages in terms of adaptation, generalization, and easy to deploy.

Although the results demonstrated strong forecast performance, this study has several limitations that should be noted. First, the test is limited to a single dataset and to one target variable (LIT301), which does not necessarily represent a wider cyber-physical system with a high diversity of operational behavior. Second, zero-shot and few-shot prompting techniques are vulnerable to changes in the input formatting, structure of demonstrations, and length of sequences. This brings about unpredictability in reproducibility between deployments. Moreover, the computational power of stronger and larger LLMs like o4-mini is still much more expensive than an architecture with edges.  This hinders real-time predictions when there is a lack of connectivity or when GPU resources are restricted. The problem of interpretability has also yet to be resolved, as pretrained foundation models do not necessarily offer clear interpretations of sensor-level patterns of decisions. Lastly, this research does not consider concept drift, persistence of attacks, actuator failures, and other dynamic nuisances that are typical in time series in the real world. Addressing these limitations will be essential to ensure reliable, secure and scalable forecasting in operational IoT infrastructures.

\subsection{Explainability of the LLMs}
In these experiments, we used Large Language Models (LLMs) in direct communication, via their respective Application Programming Interface (APIs), such as OpenAI o-4-mini and Gemini 2.5 Flash-Lite. We used structured prompts as the main communication interface. The models remained untuned for the new time series data. However, we fully relied on prompt-based conditioning and guidance on the zero-shot and few-shot prompting. In the zero-shot configuration, no labeled (i.e., a sample of next value of time series data) examples were provided in that the model was presented with just a description of the task and the instances of time series data. This setting allowed us to evaluate the model’s ability to generalize purely from instructions provided through prompts. The examples of the prompts used in our study are provided below. Each prompt includes a brief description of the task, a formatting requirement, and a single data sample.

\begin{tcolorbox}[title=Zero-shot Prompt, colback=gray!3, colframe=black!40]
    % \small
    You are given a time series window (normalized, mean 0, std 1).\\[2pt]
    \textbf{Input:}
    \texttt{[2.222027459944977, 2.222027459944977, 2.222027459944977, 2.222027459944977, 2.222027459944977, 2.222027459944977, 2.222027459944977, \dots]}\\[4pt]
    \textbf{Task:} Predict the next 1 value(s) of the series (same normalized scale).\\[2pt]
    \textbf{Output Format:} JSON object exactly as: \texttt{\{"pred": [v1, v2, ...]\}}.Return numbers only\\[2pt]
    
\end{tcolorbox}

\begin{tcolorbox}[title=LLM response to Zero-shot Prompt, colback=gray!3, colframe=black!40]
    % \small
    \textbf{LLM Response:}
    \begin{lstlisting}[basicstyle=\ttfamily\small]
        {"pred": [1.1356938766157008]}
    \end{lstlisting}
\end{tcolorbox}

 In contrast, the few-shot prompts contained two representative samples of input-output pairs, which were written in-text as part of a single prompt. Individual prompts always had 250-500 tokens based on the complexity of the example set and the characteristics of the task. For more structured tasks, the few-shot size prompt was close to the upper limit due to the involvement of domain-specific features. All calls to the LLM were made synchronously through API so that request response pairs can be deterministically logged to be evaluated. The examples of the few-shot prompts used in our experiments are shown below.

 \begin{tcolorbox}[title=Few-shot Prompt, colback=gray!3, colframe=black!40]
% \small
    You are given example windows and their next value(s) (normalized).\\[2pt]
    \textbf{Window:} \texttt{ [1061.62683 1060.90588 1060.90588  ...]} \\[2pt]
    \textbf{Next:} \texttt{[1025.8988]} \\[2pt]
    \textbf{Window:} \texttt{[1060.90588 1060.90588 1060.58545 ...]} \\[2pt]
    \textbf{Next:} \texttt{[1025.8988]} \\[2pt]
    \textbf{Input:}
    \texttt{[2.222027459944977, 2.222027459944977, 2.222027459944977, 2.222027459944977, 2.222027459944977, 2.222027459944977, 2.222027459944977, \dots]}\\[4pt]
    \textbf{Task:} Predict the next 1 value(s) of the series (same normalized scale).\\[2pt]
    \textbf{Output JSON exactly:} \texttt{\{"pred": [v1, v2, ...]\}}.Return numbers only\\[2pt]

\end{tcolorbox}

\begin{tcolorbox}[title=LLM response to Few-shot Prompt, colback=gray!3, colframe=black!40]
% \small
\textbf{LLM Response:}
    \begin{lstlisting}[basicstyle=\ttfamily\small]
    {"pred": [1.1356938766157008]}

    
    \end{lstlisting}
\end{tcolorbox}

To address explainability, the LLM generate its predictions along with the natural language explanation of each forecast. In the zero-shot setting, the model relied almost entirely on recent temporal structure in the window.The consistency of the final readings and extrapolated the short-term behavior which lead to consistent reasoning and had the highest quantitative performance.The reasons were usually point to trend detection, plateau identification and continuity of the latest timeseries value.In contrast, few-shot prompting encouraged the model to follow the trend of the given demonstrations, yet the results were quite unstable and different in rationale. Because the model tried to generalize over examples with various scales and drift patterns. In the few-shot configuration, we provided two example windows, and each window consisted of 720 values. Although few-shot prompting produced richer textual explanations. The arguments were less consistent with the actual temporal dynamics of the target window. Whereas the predictions were less accurate than those of zero-shot. In general, the explanation traces indicate that the internal reasoning of the LLM is highly localized around the latest values in the sequence. Where as zero-shot prompting produced more grounded and consistent and reliable explanations of our small-scale timeseries data.

% prompts did not ask the LLMs to provide any natural-language explanations. Rather, the model was strictly required to produce a JSON object with the predicted value(s) of the time series for the next. Consequently, all the models, regardless of the zero-shot and few-shot mode, provided the numerical prediction only in the given JSON format, without the text or reasoning. Because the models were constrained to output numbers only, no qualitative explanation or rationale was produced. Thus, no comparison of explanation quality between zero-shot and few-shot prompting can be made. The differences in the behavior of the observation alone are due to the numerical behavior. In the zero-shot setting, such prompts always produced more consistent and less variable predictions.  Conversely, in few-shot prompts there was a tendency for more and less variability depending on the example windows used. In the few-shot configuration, we provided two example windows, and each window consisted of 720 values. Even without textual rationales, these structured outputs allow us to examine the impact of prompting strategy on prediction consistency in LLM-based time series forecasting.

\section{Conclusion}
\label{sec:conclusion}

This paper conducted a systematic comparison of seven prediction strategies with a time series data set. The results provide strong evidence that pretrained time-series foundation models such as TimesFM enable superior predictive performance with minimal computational adaptation. TimesFM was the best, with the lowest RMSE of 0.3025 and MAE of 0.2127, while requiring only 266 seconds of inference time. The result of TimesFM shows that it can effectively model multi-sensor nonlinear interactions without requiring the backpropagation-based model optimization. OpenAI o4-mini (zero-shot) showed very competitive results (RMSE 0.3310, MAE 0.2098) even with no task-specific examples. This confirmed that semantic pattern reasoning in large transformer decoders can be effectively applied to time series forecasting. Nevertheless, its cost in inference was still high (more than 21k sec), which is an indication of obvious trade-offs in deployment.

In contrast, deep learning architectures had the worst forecasting ability with the same training budget. With 20 epochs of training, both LSTM and TCN achieved RMSEs of 0.7361 and 0.7174, respectively. These models lack the global temporal receptive fields and large-scale priors embedded in foundation models, limiting their ability to detect fast-evolving process disruptions.  Gemini 2.5 Flash Lite was also having problems adapting to multivariate signal behavior (RMSE $\geqslant$ 0.4060) and exhibited higher latency compared to TimesFM. The results suggest that lightweight language-reasoning alone does not automatically translate into forecasting strength for time series data.

In a deployment perspective, the findings indicate that foundation models exploit pretrained temporal abstractions to achieve greater performance than architecture-specific learning, even in noisy conditions or seasonality which impact in time series data. TimesFM is the most feasible candidate for time series forecasting and edge-deployable monitoring pipelines due to its massively trained time pretraining and its practical decoder-only network. Future work will focus on reducing the inference cost in resource-limited environments. This will include optimization techniques such as model quantization and knowledge distillation. We will also incorporate uncertainty-aware probabilistic estimates to strengthen anomaly detection. These improvements will support safer autonomous control decisions in industrial systems. They will also help defend against adversarial manipulation in critical infrastructure. In general, this study shows that time-series foundation models are essential for improving predictive reliability. They also enhance operational resilience across dynamic time series forecasting environments.

\section*{Acknowledgment}
This work is partially supported by a grant from the National Science Foundation (Award No. 2319802).

\bibliography{refs}{}
\bibliographystyle{plain}

\end{document}